\documentclass{cta-author}
\usepackage{multirow}
\usepackage{subfigure}
\usepackage{url}
\usepackage{pgf,interval}
\usepackage{csvsimple}
\usepackage{soul}
\usepackage{xcolor,colortbl}
\definecolor{pearl}{rgb}{0.94, 0.92, 0.84}
\definecolor{aliceblue}{rgb}{0.94, 0.97, 1.0}
\definecolor{lavenderblue}{rgb}{0.8, 0.8, 1.0}
\definecolor{palespringbud}{rgb}{0.93, 0.92, 0.74}
\definecolor{pearl}{rgb}{0.94, 0.92, 0.84}
\usepackage{makecell}
\usepackage{arydshln}
\usepackage{changes}
\usepackage{lipsum}

\usepackage[normalem]{ulem}
{}
{}
{}
\begin{document}
\supertitle{Submission Template for IET Research Journal Papers}
\title{Energy Forecasting in Smart Grid Systems: Recent Advancements in Probabilistic Deep Learning}
\author{\au{Devinder Kaur$^{1}$}, \au{Shama Naz Islam$^{2}$}, \au{Md Apel Mahmud$^{3\corr}$},  \au{Md Enamul Haque$^{4}$}, \au{Zhao Yang Dong$^{5}$}}
\address{~} 
\add{1,2,4}{School of Engineering, Deakin University, Waurn Ponds, VIC 3216, Australia}
\add{5}{School of Electrical Engineering and Telecommunications, The Univ of NSW, Sydney, Australia}
{\textbf{Corresponding Author:}} 

\add{3\corr}{Electrical Power and Control Systems Research Group Department of Mathematics, Physics and Electrical Engineering Northumbria University, Newcastle Upon Tyne, NE1 8ST, United Kingdom. Email: md.a.mahmud@northumbria.ac.uk}
\begin{abstract}
Energy forecasting plays a vital role in mitigating challenges in data rich smart grid (SG) systems involving various applications such as demand-side management, load shedding, and optimum dispatch. Managing efficient forecasting while ensuring the least possible prediction error is one of the main challenges posed in the grid today, considering the uncertainty in SG data. This paper presents a comprehensive and application-oriented review of state-of-the-art forecasting methods for SG systems along with recent developments in probabilistic deep learning (PDL). Traditional point forecasting methods including statistical, machine learning (ML), and deep learning (DL) are extensively investigated in terms of their applicability to energy forecasting. In addition, the significance of hybrid and data pre-processing techniques to support forecasting performance is also studied. A comparative case study using the Victorian electricity consumption in Australia and American electric power (AEP) datasets is conducted to analyze the performance of deterministic and probabilistic forecasting methods. The analysis demonstrates higher efficacy of DL methods with appropriate hyper-parameter tuning when sample sizes are larger and involve nonlinear patterns.
Furthermore, Bayesian inference integrated with bidirectional long short-term memory (BLSTM) as a PDL method 
achieves lower prediction errors in addition to quantifying uncertainties.
However, the execution time increases significantly for PDL methods due to large sample space and a trade-off between computational performance and forecasting accuracy needs to be maintained

\textbf{Keywords}: Bayesian methods, energy forecasting, machine learning, power systems, probabilistic deep learning, time-series analysis, uncertainty estimation
\end{abstract}
\maketitle

\section{Introduction}
Energy forecasting has a crucial role to play in 
planning, investment, decision making, and mitigating operational and management challenges in modern power systems
popularly termed as smart grid (SG) systems \cite{gungor2011smart}.
With the advent of smart meters and advanced metering infrastructure (AMI) in SG, a significant increase in the bidirectional flow of energy and data is observed between the grid and end-users. Thus, a number of data analytics applications such as energy forecasting have emerged recently in the SG domain. Such applications can be highly useful for scheduling generation, implementing demand response strategies, and ensuring financial benefits through optimum bidding in the energy market \cite{Josep:2020,Shama:2019}. In the past decades, traditional statistical methods such as autoregressive integrated moving average (ARIMA) and associated techniques have been extensively used for forecasting energy demand and generation \cite{agoua2017short}. However, in recent times due to the breakthrough of smart metering and thus high volume of data generation, statistical methods face scalability issues and can not analyze complex nonlinear data features \cite{jahangir2020deep}.  

Since the last decade, artificial intelligence (AI)-based methods involving machine learning (ML) and deep learning (DL) algorithms have attracted much attention for their ability to generate accurate forecasts in SG systems \cite{wang2018review}. Particularly, sequence-based DL algorithms such as recurrent neural networks (RNN) and long short-term memory (LSTM) models have proven to be potential methods to deal with the nonlinear energy data features with long sequences \cite{kaur2019smart}. 
Furthermore, SG data generation experience the problem of stochastic uncertainties when renewable generation intermittencies and variations in energy consumption behaviors are considered. 
Moreover, tackling parametric or model uncertainties is another challenge faced by aforementioned forecasting methods, which are deterministic in nature and provide point forecasts.
In such cases, probabilistic methods \cite{kaur2022bayesian,zhang2014review,liu2015probabilistic} having the ability to generate prediction intervals (PIs) highlight their effectiveness to manage uncertainties compared to point forecasting methods. Recently, probabilistic deep learning (PDL) has made its way to carry out the forecasting more efficiently by integrating deep neural networks and Bayesian inference \cite{ali2020bayesian,yang2019bayesian}. However, 
this approach needs to be investigated more and offers a substantial scope in modern power systems and forecasting applications.

Furthermore, by ensembling multiple methods as one hybrid approach, forecasting accuracy can be highly improved \cite{Torabi:2018}. However, the model complexity could be one of the main shortcomings of such approach and thus, a trade-off between the accuracy and computational complexity needs to be maintained.
In this direction, data pre-processing may play a crucial role in improving the model performance and minimizing the forecasting error \cite{kaur2018tensor}. In this context, dimensionality reduction (DR) and feature extraction are the key approaches often adopted for effective data pre-processing \cite{kaur2021vae,featureXtraction}.
In this regard, principal component analysis (PCA) \cite{huang2019optimization} and singular value decomposition (SVD) \cite{de2015data} have been widely used as underlying pre-processing techniques with different forecasting methods to deal with the challenges of high-order dimensionality in SG data. From DL domain, auto-encoders (AE) implemented with convolutional layers have been used as a feature extraction scheme to filter multiple dimensions in consumption data \cite{ryu2019convolutional}.

Although, there have been a significant number of research papers which surveyed ML methods for solar irradiance forecasting \cite{Cyril:2017,Naveed:2019,Dhivya:2020} and the authors in \cite{Wang:2019} reviewed state-of-the-art DL methods for renewable energy forecasting.  
None of the aforementioned review papers investigated advanced DL methods such as RNN and LSTM in detail for energy forecasting applications with a comparison made between the statistical, ML, and DL methods.
Furthermore, to the best of our knowledge, no review paper has yet considered reviewing the recent developments in probabilistic forecasting domain, especially contributions of probabilistic deep learning for energy forecasting applications in SG systems.
This motivates us to review state-of-the-art forecasting methods holistically along with the recent developments in PDL domain in an application-oriented manner.

To be specific, we make the following contributions:

$\bullet$ A comprehensive review of statistical, ML, DL, probabilistic, and hybrid forecasting methods is presented along with their applications in SG systems considering different time horizons. The existing pre-processing techniques to aid energy forecasting performance are also discussed.

$\bullet$ A number of statistical, ML, DL, and PDL techniques are implemented for two different energy consumption datasets with variable time resolution to compare forecasting accuracies. The impact of high variability and data sizes on different methods is also evaluated.	
	
$\bullet$ A comparative case study is presented to evaluate the performance of aforementioned forecasting methods.
The analysis shows that RNN and LSTM can achieve higher accuracies with least prediction errors under deterministic forecasting methods with larger sample sizes especially in the presence of high variability.
However, the accuracy can vary with the choice of activation function and hyper-parameters tuning which needs to be appropriately selected for a given dataset.

$\bullet$ For probabilistic forecasting, Bayesian bidirectional LSTM (BLSTM) is considered against state-of-the-art methods and Bayesian with standard artificial neural networks (ANN). From the implementation results, it can be inferred that Bayesian BLSTM outperforms deterministic methods by exhibiting least error for all data variations along with providing prediction intervals for energy forecasting scenarios. Though Bayesian DL methods can improve the forecasting accuracy, they have a higher execution time as compared to deterministic methods, and a trade-off between the model performance and computational cost is required.

The rest of the paper is organized as follows.
Section \ref{sec:applications} discusses the applications of forecasting in SG systems. Section \ref{sec:timehorizon} outlines the time horizon-based categorization of forecasting methods. Section \ref{sec:taxonomy} describes the taxonomy of the energy forecasting methods. Section \ref{sec:deterministic} elaborates the different deterministic forecasting methods. The contributions related to probabilistic forecasting methods have been described in Section \ref{sec:probabilistic}. A number of hybrid methods have been presented in Section \ref{sec:hybrid}. Further, data pre-processing methods are briefly discussed in section \ref{sec:preprocessing}. Section \ref{sec:casestudy} represents a comparative case study of energy forecasting methods on two different datasets. Finally, Section \ref{sec:conclusion} concludes the paper and discusses the future directions.
\begin{figure}
	\centering
	\includegraphics[width=0.5\textwidth]{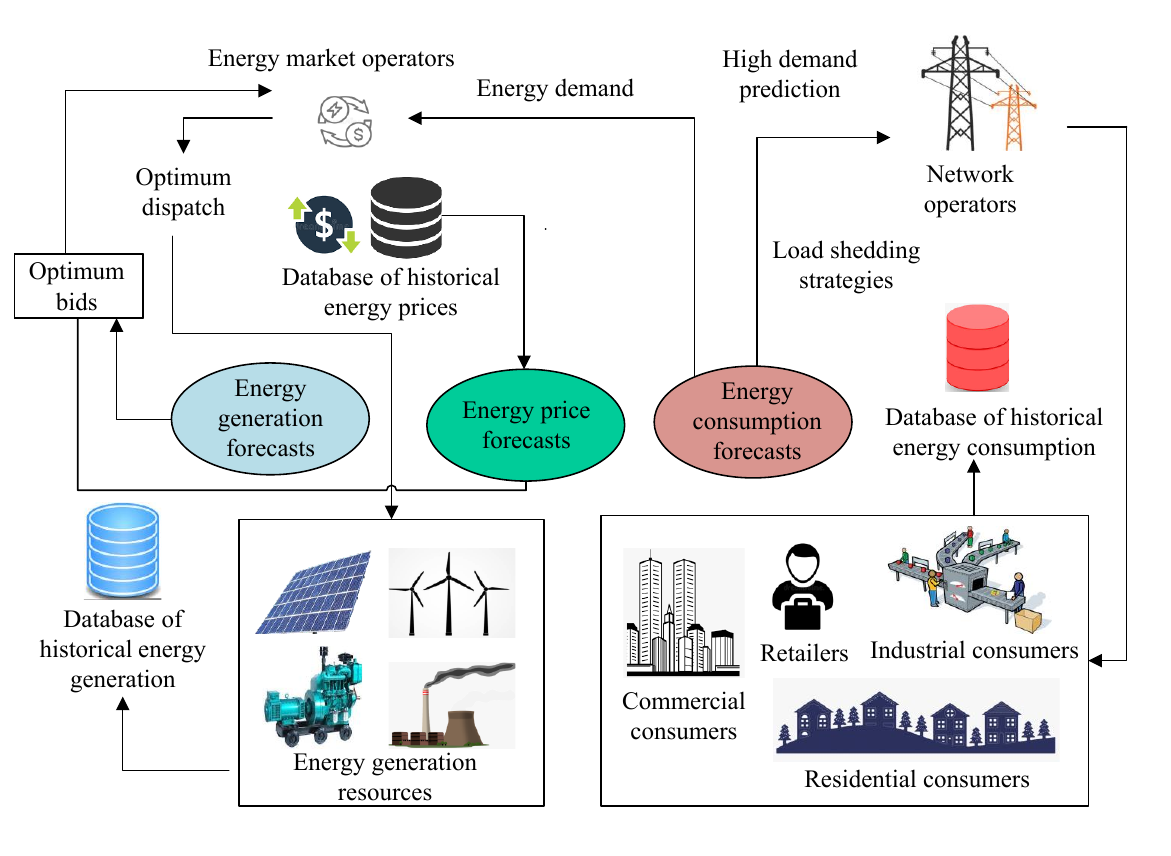}
	\caption{Overview of forecasting applications for energy systems}
	\label{overview}
\end{figure}
\section{Applications of forecasting in Energy systems}\label{sec:applications}
\indent Energy forecasting with high accuracy and precision can provide significant insights for grid planning and reliability and has a wide range of applications including load shedding, optimum dispatch, peak load shaving, etc. \cite{hernandez2014survey}. The data generated from SG systems is usually related to the energy consumption by end users involving residential, commercial, and industrial consumers as shown in Fig. \ref{overview}. Furthermore, additional data is generated from distributed energy resources (DERs) such as solar and wind power plants. The generated data is further utilized by forecasting methods to forecast energy generation, demand, and prices in SG systems. The authors in \cite{rossi2019smart} surveyed various challenges and trends of demand forecasting with respect to different time horizons and regions. Similarly, in \cite{hsiao2014household} authors compared existing statistical methods for demand forecasting and emphasized the significance of attaining least root- mean-square error (RMSE). Further, Kong \textit{et. al.} in \cite{kong2017short} employed DL methods to forecast short-term energy consumption and demand for individual households using highly granular data. 

As energy generation using renewable-based DER is more intermittent due to exogenous factors such as weather changes and user behaviors, ensuring high forecasting accuracy is a very challenging task in the grid today \cite{voyant2018prediction}. In this regard, \cite{liu2018vector} used vector autoregressive (VAR) to forecast solar irradiance, temperature, and wind speed for 61 locations in the United States. Similarly, Messner \textit{et. al.} in \cite{messner2019online} used VAR method to forecast wind power generation based on high dimensional data.

Furthermore, 
forecasting price spike is an important application which considers the biggest risk factor in the energy market. Recent zero or negative pricing as observed in the Australian national electricity market (NEM) can negatively impact generators \cite{pino2008forecasting}. On the other hand, positive high price spikes can give more profits to generators, especially during peak generation hours \cite{ zhao2007framework}. Yang \textit{et. al.} surveyed the latest trends in the decision making for electricity retailers using consumed load price forecasting \cite{yang2017decision}. Furthermore, \cite{Miltiadis:2015} utilized a set of relevance vector machines to predict the price for individual time periods ahead-of-time and implemented micro-genetic algorithm to optimize linear regression ensemble coefficients for aggregated price forecasts. Toubeau \textit{et. al.} in \cite{toubeau2018deep} utilized PDL method to forecast wind and PV generation in return to predict the electricity prices generated from renewable DERs. Moreover, the locational marginal price for optimal scheduling of the energy storages is performed using artificial neural networks (ANN) in \cite{Meng:2014}.
\section{Time horizons for Energy Forecasting methods}\label{sec:timehorizon}
The time period over which forecasts are generated is defined as the time horizon or timescale. It is one of the key parameters to categorize forecasting approaches in SG systems \cite{soman2010review}.
Depending on time horizon, energy forecasting can be broadly classified as following: 

$\bullet$ Very short-term forecasting (VSTF): This class of forecasting involves time horizon from minutes to few hours, usually between ($0$-$3$ h) \cite{book}. VSTF can help dealing with random changes in renewable energy generation which can be predicted only before a short period of time. It offers a wide range of applications in renewable energy resources (RES) such as wind and solar power forecasting \cite{guan2012very,guan2013hybrid}. In this regard, Potter \textit{et. al.} in \cite{potter2006very} presented a $2.5$ min ahead forecasting system for Tasmanian wind farms using ANN and fuzzy logic as a hybrid approach.
	
$\bullet$ Short-term forecasting (STF): It involves energy forecasts ranging from few minutes to a few days ahead. This class plays a prime role in various grid operations such as dispatch scheduling, reliability analysis, etc. \cite{lopez2018parsimonious}. Furthermore, accurate STF helps avoiding underestimation and overestimation of the demand and thus, substantially contributes to the grid reliability \cite{fan2011short, ribeiro2019enhanced}.
	
$\bullet$ Medium-term forecasting (MTF): It implies to horizons ranging from few days to a few months ahead within a year \cite{amjady2010midterm}. MTF supports adequacy assessment, maintenance, and fuel supply scheduling in SG systems. Furthermore, it contributes to risk management using price forecasting and therefore, plays a significant role in evaluating the financial aspects of energy systems \cite{wang2017optimal}.
	
$\bullet$ Long-term forecasting (LTF): It involves horizons measured in months, quarters, and even years. LTF is crucial for load growth and energy generation planning operations over longer periods of time \cite{hong2013long,xu2017long}. 
LTF helps removing the impact of random fluctuations arising in shorter term and predict the longer term trends.
In this context, Azad \textit{et. al.} \cite{azad2014long} predicted the wind speed trends of two meteorological stations in Malaysia for one year using neural networks to manage the challenges posed by intermittent nature of wind generation. 
\begin{figure}
	\centering
	\includegraphics[width=0.45\textwidth]{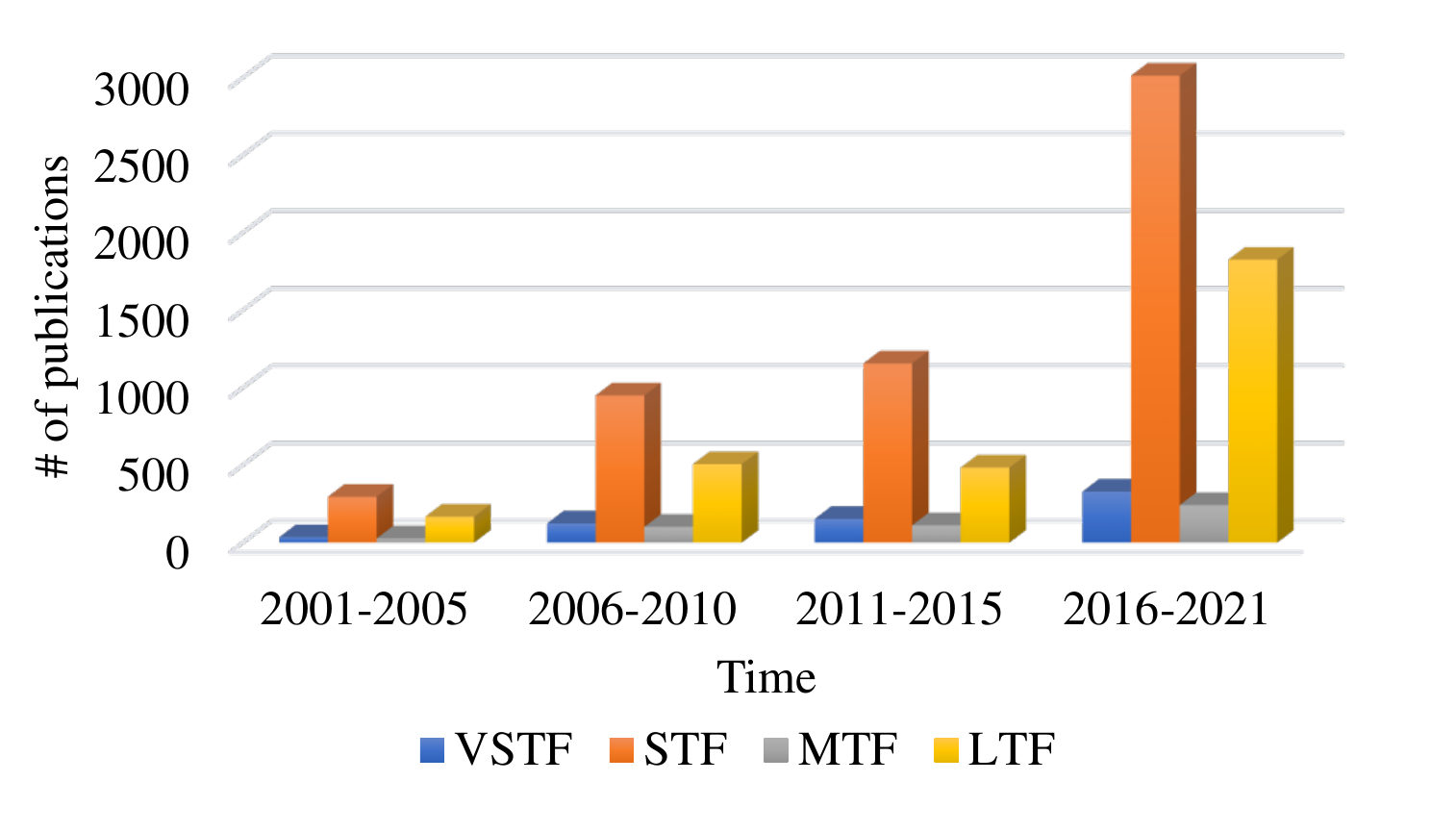}
	\caption{Publication trends in energy systems with respect to time horizons}
	\label{publications}
\end{figure}

Fig. \ref{publications} shows the pattern of publications for last two decades within five year duration with respect to different time horizons in energy systems forecasting. While LTF stands second in line, most number of publications are made for STF in the period 2016-2021, making it most widely utilized forecasting category in recent times for different applications in grid planning, operations, and energy management.
Appropriate selection of the forecasting horizon is crucial for model training and hyper-parameter tuning of a forecasting method.  
In this context, the authors in \cite{Cyril:2017} and \cite{Naveed:2019}
emphasized to focus on the selection of relevant time horizons while building the ML model for solar irradiation forecasting.  
\section{Taxonomy of the energy forecasting techniques}\label{sec:taxonomy}
This section discusses how we have categorized the state-of-the-art methods and recent advancements in energy forecasting systems along with their literature and applications. We have utilized different criteria to categorize energy forecasting methods including time horizon, representation of the forecasting output (point vs distribution), and the model performance (using error metrics) used to generate the forecasts. Given that any forecasting application can be quantified based on how far in the time horizon the output can be forecast, how much the forecasts can vary or how the forecasting model learns, the defined criteria can sufficiently capture the different dimensions of forecasting problems.

\begin{figure*}[ht!]
	\centering
	\includegraphics[height=0.425\textheight,width=\textwidth]{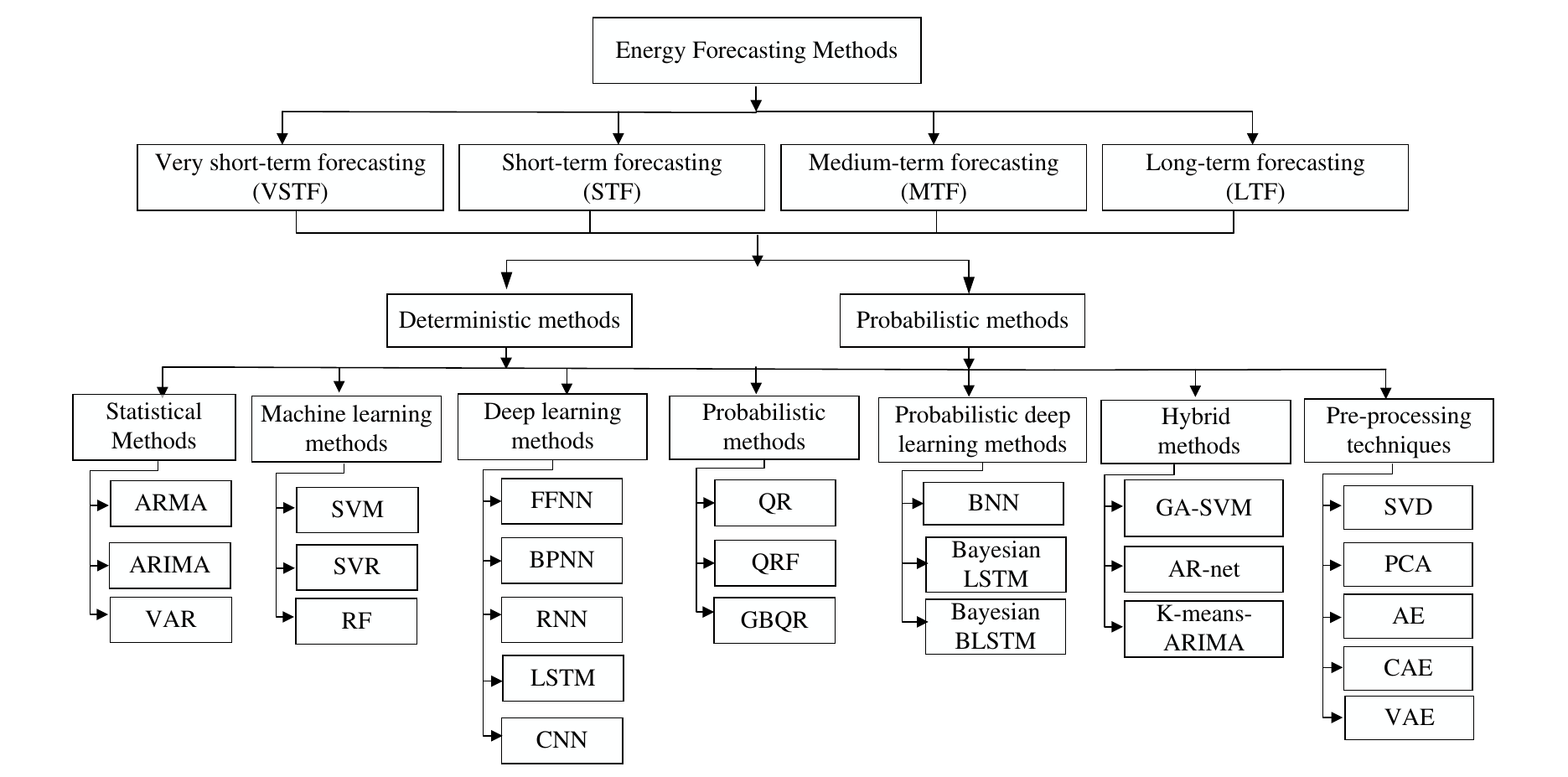}
	\caption{Taxonomy of energy forecasting methods for smart grid systems}
	\label{flow}
\end{figure*}

According to the defined criteria, energy forecasting methods can be broadly classified into VSTF, STF, MTF and LTF, as explained in the previous section. In terms of the forecasting model output type, the methods can be grouped into deterministic and probabilistic methods. Deterministic methods focus on generating point forecasts, whereas probabilistic methods generate forecasts in terms of prediction intervals. In terms of how the forecasting models are trained to predict future demand and generation, forecasting methods can be categorized as statistical, artificial intelligence (AI)-based methods (such as ML and DL), quantile regression-based methods, and the recent probabilistic deep learning (PDL) methods, as illustrated in Fig. \ref{flow}. While statistical methods are traditional and simpler in form, learning based methods involve ML, DL and its variants which are considered to be more accurate but complex at the same time.
Note that, the classification of these methods may have some cross-overs among different groups. For example, statistical methods have been explained under deterministic techniques, but these can also be formulated as probabilistic methods, and have been discussed under probabilistic techniques. Similarly, ML and DL methods can also be combined with probabilistic methods and have been covered under PDL methods.
Furthermore, hybrid methods and data pre-processing techniques have been categorized under the same two categories as these techniques can be utilized in either way, deterministic or probabilistic. 

The following sections discuss state-of-the-art and advanced forecasting methods in detail with their literature and applications for SG systems in accordance with the categorization reflected in Fig. \ref{flow}. 
\section{Deterministic techniques}\label{sec:deterministic}
Deterministic techniques generate forecasts in terms of fixed points rather than a probability distribution. These methods are also termed as point forecasting methods.
\subsection{Statistical methods}
Statistical methods are classical in nature, widely used for time-series forecasting especially for STF using the historical SG data \cite{deb2017review}. These methods are based on fitting a regression model on historical data points and then validating the model by finding the difference between actual and predicted values. This subsection stresses about existing statistical methods for the applications of SG systems forecasting.
\subsubsection{Autoregressive moving average model (ARMA)}
ARMA is the fundamental statistical method widely employed for time-series analysis. It is the combination of autoregressive (AR) and moving average (MA) methods. The AR part predicts the value at a time stamp $t$ as a function of its previous value at $t-p$, where $p$ is the order of AR model. The MA part linearly combines the error term with the previous values to form the observed data.
Sansa \textit{et. al.} in \cite{sansa2020solar} utilized ARMA to predict the solar irradiation for a winter day with maximum $10\%$ variation in the generation.
However, ARMA can only be applied to a stationary time-series, for which mean and variance do not change over time and has an uniform distribution.
\subsubsection{Autoregressive integrated moving average (ARIMA)}
To deal with the non-stationary time-series, ARMA is generalized to ARIMA. The integrated component deals with the non-stationary time-series by replacing the data values with differences of its current and previous values. ARIMA is employed in \cite{arima6} for energy demand forecasting to charge the electric vehicles (EVs) using historical load data. Similarly, it has been used to forecast energy usage in \cite{ARIMA2} and \cite{ARIMA3} for control and optimization in residential microgrids. 
However, the ARIMA is more suitable for linear time-series as it gives relatively high RMSE values for nonlinear data and requires a high execution time for large datasets \cite{kedrowski2018short}. To manage non linearity, ARIMA models often have to utilize additional techniques such as log and box-cox transformations.
\subsubsection{Vector autoregression (VAR)}
VAR model is defined as an extension to the univariate autoregression for $n$ number of time-series.
To capture linear dependencies among multiple time-series, VAR has been substantially adapted as another statistical method in the applications of energy systems forecasting. 
The authors in  \cite{liu2018vector} predicted temperature, solar irradiation, and wind speed using VAR for 61 locations in the United States and exhibited lower RMSE values as compared to other statistical methods such as simple persistence (SP). 
VAR is also used in \cite{messner2019online} for short-term wind power forecasting to tackle the problem of estimation by integrating lasso estimator to recursively update the VAR parameters.  
\subsubsection{Autoregressive/Autoregressive integrated moving average with exogenous inputs \\ (ARMAX/ARIMAX)}
ARMAX/ARIMAX models take into account external variables that influence the forecasting accuracy of a desired parameter. It combines the AR and MA processes (along with the integrated component for ARIMA) with the external time-series parameters \cite{LI201478}. The authors in \cite{LI201478} utilized temperature, humidity, irradiance duration parameters as exogenous inputs to improve the accuracy of the solar generation forecasting through an ARMA model and demonstrated that the ARMAX model achieves better accuracy in comparison to standard ARIMA models. On the other hand, the authors in \cite{MARCHESONIACLAND20201140} have employed satellite images as exogenous variables with ARMA model integrated with recursive least squares filter for intra-day solar irradiation forecasting. Energy forecasting applications often face challenges from seasonality aspects. For example, user load demand can vary daily, weekly, and annually, which should be accounted for when considering electricity consumption forecasting. In this essence, the authors in \cite{ELAMIN2018257} have developed a seasonal ARIMA framework with weather variables and seasonality as exogenous inputs and the correlation of energy consumption with weather parameters as interaction inputs for forecasting load demand. Though the aforementioned methods achieve good forecasting accuracy, the ML methods that are discussed in the following subsection are found to be more efficient and scalable to deal with nonlinear traits of SG datasets with multiple time-series \cite{ANNvsARIMA}. 
\subsection{Machine learning (ML) methods}
ML is a set of computing algorithms to learn patterns from input data and improve the prediction accuracy automatically over time.
It has a wide range of applications in big data analytics, data mining, and computer vision \cite{wang2018review}. ML is broadly categorized into supervised and unsupervised learning \cite{ma2019review}. The former involves labeled data and its two main classes are classification and regression. On the other hand, unsupervised learning deals with the unstructured data and works on grouping the data points based on their similarities and correlation \cite{kotsiantis2007supervised}.
Following subsections discuss existing ML methods and their applications for energy systems forecasting. In \cite{ma2019review}, Ma \textit{et. al.} reviewed various ML algorithms in power systems and emphasized their effectiveness for short-term load forecasting in terms of accuracy and scalability. Support vector machine (SVM), support vector regression (SVR), and random forest (RF) are the main ML methods discussed under supervised learning in the following subsections for energy forecasting applications.
\subsubsection{Support vector machine (SVM)}
SVM is widely used in ML to classify data variables into different planes with respect to margin between the data classes. The objective of SVM algorithm is to maximize the margin distance between the data variables so as to increase the accuracy of data classification. For SG systems,
Shi \textit{et. al.} proposed SVM to forecast one day ahead photovoltaic (PV) generation in China by classifying weather data into cloudy, clear, foggy, and rainy categories \cite{shi2012forecasting}. The SVM-based proposed technique demonstrated effective results with 8.64\% mean relative error (MRE). Similarly, authors in \cite{wang2018deep} employed SVM to predict socio-demographic features of users from energy consumption data with the help of convolutional filters.
\subsubsection{Support vector regression (SVR)}
The concept of SVM adapted to regression analysis is known as SVR. It supports nonlinear traits in the prediction model and provides accurate forecasting results.
Since time-series forecasting involves regression operation, SVR is comprehensively used in the existing research for demand and generation forecasting \cite{hu2015short,yang2014weather,jiang2016short,ren2014novel}. 
The authors in \cite{kavousi2014new} applied SVR for short-term load forecasting  and compared the results with other competitive techniques such as ARMA and artificial neural networks. Authors claimed that SVR integrated with modified firefly algorithm (MFA) optimizes SVR parameters and reflects better accuracy than the aforementioned methods.
Furthermore, \cite{ghelardoni2013energy} predicted long-term energy consumption using SVR after decomposing time-series consumption data into two components to justify trend and oscillations, respectively. 
\subsubsection{Random Forest (RF)}
RF is an another ML algorithm which operates as an ensemble of decision trees and often utilized for classification and regression tasks. Saeedi \textit{et. al.} implemented different ML methods including decision tree, linear regression, and multilinear perceptron (MLP) to estimate PV generation output in Hawaii \cite{saeedi2021adaptive}. They claimed the superior performance by RF with highest R-squared value of $98$\% as compared to aforementioned ML models. Additionally, weather data and location of PV were also considered while modeling the generation scenarios.
Similarly, \cite{liu2021hybrid} employed a hybrid of fuzzy c-means clustering and RF to predict short-term load consumption after grouping the similar load profiles to reduce uncertainty in the individual consumption data. Implementation results claimed high performance in terms of accuracy by proposed RF-based framework as compared to other ML methods.

However, energy datasets with high volume and complex patterns require extensive learning and optimization, where ML methods usually fail to perform efficiently \cite{shi2017deep}.
In this context, DL methods based on artificial neural networks are emerging as a highly competent set of techniques to deal with high dimensionality and uncertainty traits of SG data \cite{sun2019using}. 
\subsection{Deep learning (DL) methods}
DL methods involve artificial neural networks (ANNs) defined as sequence of layers having $n$ number of connected nodes termed as neurons \cite{li2020bibliometric}. 
An input from external data is fed to the network via input layer and an output is obtained by optimizing the network with learning algorithm via output layer. An ANN may have zero or higher number of intermediate layers known as hidden layers.
ANNs with more than one hidden layer are known as deep neural networks (DNN). 
DL algorithms have set a benchmark in the field of image and audio processing and are receiving an increased attention in SG systems, since the last decade \cite{zhang2018review,fang2019performance}.
This subsection discusses state-of-the-art DL methods employed in energy systems forecasting along with relevant literature and applications.
\subsubsection{Feed forward neural networks (FFNN)}
Vanilla ANNs without having loops within layers are known as FFNNs. They can be single or multilayer perceptrons and are primarily used for supervised learning.
Bhaskar \textit{et. al.} presented a FFNN-based wind power forecasting framework integrated with wavelet decomposition to forecast $30$ hour ahead wind power built on wind speed estimation \cite{bhaskar2012awnn}. Authors demonstrated effective results of the proposed method in comparison with persistence and non-reference benchmark methods.
Similarly, \cite{pindoriya2008adaptive} investigated the use of FFNNs for short-term price forecasting for Spanish and PJM electricity markets and conducted a comparative analysis with wavelet-ARIMA and fuzzy neural networks. Although, the standard FFNN outperformed comparative methods, they stated that FFNNs integrated with adaptive wavelet neural networks provide more satisfactory predictions by demonstrating least RMSE values.
\subsubsection{Back-propagation neural networks (BPNN)}
In contrast to FFNNs, BPNNs involve a back propagation algorithm such stochastic gradient descent (SGD) to fine tune the weight parameters at each neural layer based on the error generated by the loss function \cite{jiao2018model}. A BPNN-based ANN ensure a lower error rates and thus, more reliability in the predictions. BPNN is often integrated with optimization techniques or other ML models for improved performance. For example, wind speed prediction has been performed in \cite{Yagang:2018} by optimizing the weight parameters of BPNN using improved particle swarm optimization. On the other hand, a flower pollination algorithm has been implemented in \cite{Zongxi:2019} to optimize the BPNN for wind speed forecasting. In the context of energy consumption forecasting based on socio-econometric factors, a differential evolution technique integrated with adaptive mutation is implemented to optimize the initial connection weights of BPNN \cite{Yu-Rong:2017}. Following subsections involve neural networks based on underlying BPNN algorithm.
\subsubsection{Recurrent neural networks (RNN)}
ANNs with directed cycles between hidden layers are defined as RNN. Contrary to FFNNs, they have a memory state and can process variable data length, thus facilitating the sequence processing in time-series.
Shi \textit{et. al.} proposed RNN for STF of granular consumption data for $920$ households in Ireland \cite{shi2017deep}.
However, RNNs face the problem of over-fitting and may reflect false positive results for new data values. They addressed this issue by increasing diversity and volume in individual consumption data by introducing a pooling system of neighboring households with correlated historical consumption. 
According to the results, RNN with pooling technique outperforms benchmark forecasting techniques such as ARIMA, SVR, and classical RNN by significantly reducing RMSE values. Similarly, RNN is used in \cite{kaur2019smart} to predict short-term energy consumption values for $112$ households and promising results in the term of lower RMSE and MAE error values are reflected.
However, RNN suffers from the problem of vanishing and exploding gradients during the learning process with data having long sequences.
\subsubsection{Long short-term memory (LSTM)}
LSTM is a variant of RNN which has made a breakthrough in well-known DL applications such as natural language processing (NLP) by dealing with vanishing/exploding gradients through its cell and gating architecture. In this direction, authors in \cite{kong2017short} proposed a RNN-based LSTM technique to perform short-term load forecasting for residential consumers and demonstrated improved accuracy at granular and aggregated level forecasts. 
In \cite{motepe2019improving}, authors included neuro-fuzzy logic in conjunction with the LSTM and found that the technique can outperform other methods. The authors in \cite{zhang2020data} combined persistence model for predicting sunny weather and utilized this information to improve the solar power forecasting using LSTM integrated with auto-encoders. In \cite{shao2020multi}, authors combined convolutional auto-encoder and K-means algorithm to extract features related to customer behavior and integrated them with LSTM models for short-term energy consumption forecasting.
\subsubsection{Convolutional neural networks (CNN)}
CNN is a class of deep neural networks, extensively used in feature extraction and data filtering applications.
In \cite{wang2017deterministic}, CNN is used in conjunction with wavelength transform (WT) method to extract features from PV generation dataset to support effective forecasting. The proposed method is tested on real PV dataset acquired in Belgium and 
demonstrates superior results with improved forecasting accuracy as compared to the conventional methods.
Furthermore, Lee \textit{et. al.} investigated CNN filters integrated with LSTM to forecast one day-ahead PV solar power in Korea, considering the local weather information alongside \cite{cnnlstm}. They emphasized the need to pre-process and refine the input data to achieve high forecasting performance and compared their method with other ML techniques for different time horizons within a day and observed lower RMSE and mean-absolute percentage error (MAPE) values.

Authors in \cite{almalaq2017review} and \cite{zhang2018review} have reviewed various DL-based energy forecasting methods and conducted a case study to compare the accuracy of different methods in terms of RMSE and mean absolute error (MAE). The addition of more layers to a neural network can increase the forecasting accuracy. However, it sometimes overfits the model to a specific dataset as it captures the noise during training phase. 
To overcome this challenge, dropout \cite{srivastava2014dropout} was proposed as a potential technique which addresses overfitting by dropping out the random neuron units. 
However, uncertainty representation in generation data from RES is considered as another challenging task for point forecasting methods discussed in above sections. The uncertainty generally arises due to external factors such as weather conditions and cannot be captured very well by point forecasting methods.
\begin{table*}[!t]
	\centering
	\renewcommand{\arraystretch}{3.0}
	\caption{Existing state-of-the-art forecasting methods: A comparative summary}
		\begin{tabular}{p{5mm}p{35mm}p{32mm}p{35mm}p{35mm}}
			\hline
			Sr. no. & Methods &References&Advantages& Disadvantages \\
			\hline
			\hline
		M1 & Statistical &\cite{liu2018vector}\cite{messner2019online}\cite{lopez2018parsimonious}\cite{deb2017review}\cite{arima6}	
			\cite{ARIMA2} \cite{ARIMA3}\cite{kedrowski2018short} & Simple and low computational cost& Not reliable for big data and nonlinear data components in energy datasets \\
			M2 & Machine learning (ML) & \cite{Naveed:2019}\cite{Dhivya:2020}\cite{Soheil:2020}\cite{longterm}\cite{ma2019review}
			\cite{shi2012forecasting}\cite{kavousi2014new}\cite{zhu2016short}\cite{percy2018residential}\cite{hu2015short}
			\cite{yang2014weather}\cite{jiang2016short}\cite{ren2014novel} \cite{saeedi2021adaptive}\cite{liu2021hybrid}& Efficient and capable to deal with big energy datasets & Not reliable for heterogeneous and long sequence data problems, generate point forecasts\\
			M3 & Deep learning (DL) & \cite{kaur2019smart}\cite{jahangir2020deep}\cite{Wang:2019}\cite{kong2017short}\cite{pino2008forecasting}
			\cite{azad2014long}\cite{guan2012very}\cite{wang2018deep}\cite{shi2017deep}\cite{li2020bibliometric}
			\cite{zhang2018review}\cite{fang2019performance}\cite{bhaskar2012awnn}\cite{pindoriya2008adaptive}\cite{almalaq2017review}
			\cite{srivastava2014dropout}\cite{wang2020recent} & Efficient for long and nonlinear data patterns & Overfitting, require appropriate hyper-parameter tuning, generate point forecasts \\
		M4 &	Probabilistic methods &  \cite{fan2011short}\cite{taieb2016forecasting}\cite{dowell2015very}\cite{hong2016probabilistic}\cite{hong2013long}
			\cite{zhang2014review}\cite{zhang2014review}\cite{gneiting2014probabilistic}\cite{liu2015probabilistic}\cite{wang2018combining}
			\cite{aprillia2020statistical}& Provide prediction intervals 
			(PIs) &  High computational complexity\\
		M5 & Probabilistic deep learning (PDL) &\cite{toubeau2018deep}\cite{quantile}\cite{sun2019using}\cite{ali2020bayesian}\cite{yang2019bayesian}
			\cite{gal2016dropout}& Provide PIs, reliable, and efficient & Limited literature and high computational cost\\
		M6 &	Hybrid (ensemble) &  \cite{Torabi:2018}\cite{Naveed:2019}\cite{amjady2010midterm}\cite{ARIMAann}\cite{arnet}
			\cite{hybrid}\cite{zhang2017robust}\cite{zhang2013short} & High accuracy and scalable & Application specific, high computational complexity\\
		M7 &	Pre-processing techniques & \cite{featureXtraction}\cite{ryu2019convolutional}\cite{chen2016framework}\cite{wang2016sparse}\cite{Dairi:2020}
			\cite{pereira2018unsupervised}\cite{bachhav2019latent}\cite{wang2020variational}& Improve forecasting performance & Application specific \\
			\hline
		\end{tabular}
	\label{summary}
\end{table*}
\begin{table*}[!t]
	\centering
	\renewcommand{\arraystretch}{1.5}
	\caption{Energy forecasting surveys from past 5 years}	
	M1: Statistical; M2: ML;  M3:DL; M4: Probabilistic; M5:PDL; M6: Hybrid; M7: Pre-processing
	\label{surveys}
	\begin{tabular}{p{15mm}p{25mm}p{10mm}p{10mm}p{10mm}p{10mm}p{10mm}p{10mm}p{10mm}}
		\hline	
		Publication Year & References&M1&M2&M3&M4&M5&M6&M7\\
		\hline
		\hline
		2016&\cite{nagaraja2016survey}&$\checkmark$&$\checkmark$&$\checkmark$&$\times$&$\times$&$\times$&$\times$\\
		2017&\cite{almalaq2017review}&$\checkmark$&$\checkmark$&$\checkmark$&$\times$&$\times$&$\times$&$\times$\\
		2018&\cite{kabir2018neural}&$\times$&$\times$&$\checkmark$&$\checkmark$&$\times$&$\times$&$\times$\\
		2019&\cite{wang2018review}&$\checkmark$&$\checkmark$&$\checkmark$&$\checkmark$&$\times$&$\times$&$\times$\\
		2020&\cite{quan2019survey}&$\times$&$\times$&$\checkmark$&$\checkmark$&$\checkmark$&$\times$&$\times$\\
		2020&\cite{Tao:2020}&$\times$&$\checkmark$&$\checkmark$&$\checkmark$&$\times$&$\checkmark$&$\times$\\
		2021&\cite{ASLAM2021110992}&$\times$&$\checkmark$&$\checkmark$&$\checkmark$&$\times$&$\checkmark$&$\checkmark$\\
		2022&\cite{AHMAD2022112128}&$\checkmark$&$\checkmark$&$\checkmark$&$\checkmark$&$\times$&$\checkmark$&$\checkmark$\\
		2022&Proposed survey&$\checkmark$&$\checkmark$&$\checkmark$&$\checkmark$&$\checkmark$&$\checkmark$&$\checkmark$\\
		\hline
	\end{tabular}
\end{table*}

Table \ref{summary} presents a comparative summary of aforementioned forecasting methods along with relevant references, advantages, and disadvantages of each category. Furthermore,
Table \ref{surveys} reflects the surveys published in the last five years with respect to number of methods reviewed in each. It can be observed that no survey to the best of our knowledge has attempted to review all categories of methods comprehensively. We attempt to propose a more comprehensive review especially including probabilistic approaches.
\section{Probabilistic techniques}\label{sec:probabilistic}
With the integration of RES in the modern power grid, forecasting trends are shifting from point to probabilistic in regards to the future demand and generation at disaggregated levels \cite{gneiting2014probabilistic}.
Hong \textit{et. al.} presented a review for probabilistic methods and emphasized their importance over point forecasting with ever changing needs of power industry \cite{hong2016probabilistic,hong2010short}.
\subsection{Parametric vs non-parametric approaches}
This subsection identifies the existing literature for probabilistic energy forecasting, which is mainly categorized under parametric and non-parametric approaches. The authors in \cite{zhang2014review} provided a brief review of these two approaches for wind generation forecasting.

Parametric approaches assume a certain probability density function for the parameter distribution, such as normal distribution. Dowell \textit{et. al.} proposed a parametric probabilistic scheme based on Bayesian probability and sparse VAR to forecast very short-term wind power generation in Southeastern Australian wind farms with a $5$ min interval \cite{dowell2015very}. The authors confirmed that their method achieves least RMSE in comparison to the standard AR and VAR methods. They further utilized the similar parametric approach in \cite{hong2013long} to forecast long-term probabilistic horizons for load consumption. A hybrid probabilistic deterministic approach for wind generation forecasting has been developed in \cite{Yan:2016}, where temporally local Gaussian processes are used to investigate forecasting errors.
Furthermore, for the application of price forecasting, probabilistic methods have been utilized by various authors and the relevant contributions are outlined in \cite{nowotarski2015computing,maciejowska2016probabilistic,weron2014electricity}. Though this approach simplifies the analysis and reduces computation cost, sometimes the parameter distribution may not accurately fit with a known function \cite{zhang2014review}.

On the other hand, non-parametric approaches do not assume a fixed function for the probability density of the output parameters. In this approach, the predictive probability densities of the parameter are represented by a range of quantile forecasts \cite{zhang2014review}. For wind power forecasting in the presence uncertainties, the authors in \cite{Ricardo:2012} implemented a kernel density estimation model and represented the wind power with a number of kernel functions. An ensemble-based probabilistic forecasting model was developed in \cite{Pierre:2009} which transforms meteorological data to wind power output and generates predictive distributions in a non-parametric manner. On the other hand, a non-parametric approach is considered in \cite{Khosravi2013849} to obtain prediction intervals as outputs of neural network models for generating wind power forecasts.
\subsection{Quantile regression}
Quantile regression is an extended version of linear regression and aims to obtain predictive quantiles \cite{zhang2014review}.
Taieb \textit{et. al.} in \cite{taieb2016forecasting} emphasized the need 
to predict the future demand in probabilistic format, which can further support the grid operations related to the energy generation and distribution. They employed additive quantile regression (QR) to forecast the individual energy consumption for $3696$ smart houses in Ireland with $30$ min interval.
To support probabilistic load forecasting, Liu \textit{et. al.} proposed quantile regression averaging (QRA) to obtain PIs for consumption with $90\%$ percentiles \cite{liu2015probabilistic}. The authors claimed effective forecasts using probabilistic evaluation metrics such as Pinball and Winkler score.

However, the existing probabilistic approaches in energy forecasting often suffer high computational complexities and thus, more efficient methods need to be developed and reviewed considering the current energy market scenarios \cite{wang2018combining,aprillia2020statistical}.
\subsection{Probabilistic deep learning (PDL)}
Bayesian probability incorporated with DL methods can be used to provide forecasting results in the form of PIs, contrary to traditional deep neural networks that are deterministic in nature and generate point forecasts. PDL expresses model parameters as a function of probability distributions such as normal distribution. To be precise, PDL models can predict future PIs with different percentiles that can explain the certain or uncertain factors in the energy data and hence, enables better decision making.
This section outlines the significant contributions in the field of PDL to support the applications of demand, generation, and price forecasting in modern power systems. 
\subsubsection{Bayesian neural networks (BNN)}
The concept of Bayesian probability incorporated with ANN is defined as BNN.
Yang \textit{et. al.} in \cite{yang2019bayesian} proposed BNN technique to forecast individual energy demands at household level after quantifying the shared uncertainties between different customer groups. In addition, a clustering-based data pooling system is presented to tackle the issue of over-fitting by increasing the data volume and diversity. The authors demonstrated lower Winkler and Pinball scores for probabilistic methods. Furthermore, a comparative study is presented with the benchmark point forecasting and probabilistic techniques such as QRA and quantile regression factoring (QRF). 
\subsubsection{Bayesian LSTM}
Sun \textit{et. al.} proposed a BNN integrated LSTM approach to curb the challenges posed by weather uncertainty in distributed PV generators and thereafter generated the net-load forecasts in the form of PIs with greater accuracies \cite{sun2019using}. 
In addition, they improved the forecasting performance by clustering individual sub profiles based on similar energy consumption patterns prior to applying Bayesian approach. 
They implemented their method on a real SG dataset acquired from AusGrid involving rooftop PV generation measurements for every half an hour for a period of three years. 
\subsubsection{Bayesian BLSTM}
In a similar manner, authors in \cite{toubeau2018deep} proposed a PDL technique to deal with the problem of uncertainty in energy markets. The authors integrated LSTM with bidirectional RNN by enabling the propagation of training sequence forwards and backwards and proposed their method as bidirectional-LSTM (BLSTM). The proposed network is then trained to generate a Gaussian or a non-parametric predictive distribution of the dependent variables present in the energy data such as PV generation.
Furthermore, Copula-based sampling is employed over predicted distributions to generate predictive scenarios. 
However, probabilistic methods are computationally expensive due to large sample space. 
In this regard, \cite{gal2016dropout} proposed dropout as one of the potential solutions which works as an
approximator to make the Bayesian inference process less complex, computationally. 
However, the computational complexity of the Bayesian approach remains one of the main concerns and therefore, more generalized and effective solutions need to be explored in the future research.
\section{Hybrid methods}\label{sec:hybrid}
Statistical methods face challenges from their inability to process big data, while AI-based methods have other shortcomings in terms of model complexity and dependence on large training datasets. Furthermore, deep neural networks struggle with overfitting, vanishing/exploding gradients, and appropriate hyper-parameter tuning. Therefore, a hybrid approach involving two or more methods could be more useful to overcome aforementioned limitations of state-of-the-art point forecasting methods \cite{hajirahimi2019hybrid}. In this direction, different learning and statistical methods can be integrated together along with optimization techniques to work as one holistic approach. In this context, authors in \cite{Naveed:2019} presented a comparative analysis of various hybrid techniques by integrating ML methods with optimization algorithms such as genetic algorithm (GA) to enhance the model efficacy for PV generation forecasting. Various hybrid methods along with their literature work for energy forecasting are discussed as below.
\subsection{GA-SVM}
In \cite{hybrid}, authors combined four different state-of-the-art forecasting methods namely, ARIMA, SVM, ANN, and adaptive neurofuzzy inference system (ANFIS) with GA optimization technique to forecast one hour-ahead PV power generation by utilizing green energy office building data in Malaysia. Additionally, authors considered the solar irradiance, air, and the historical solar power data as the input features. They claimed high precision in the prediction outputs with minimum error of $5.64\%$ with hybrid approach as compared to individual aforementioned methods.
\subsection{AR-net} 
As classical methods lack long-range dependencies (LRD) and neural networks are complex and lack interpretability, authors in \cite{arnet} presented a simple yet efficient technique in which AR process is modeled using neural networks and named as AR-net. To be precise, AR-net mimics the traditional AR process with the FFNNs. 
Furthermore, the AR coefficients which are trained using least square method in existing approaches, are fitted using stochastic SGD in AR-net.

The parameters of first layer in AR-net are taken as equivalent to the classical AR-coefficients. In addition, authors discussed the equivalences between FFNNs and classic AR models to overcome scalability issues while ensuring model interpretability and simplicity. They emphasized the use of deep neural networks such as RNN and CNN to deal with scalability issues. But DL models are highly complex to interpret and understand the dynamics of models and need appropriate consideration of the suitability in each time-series application.
\subsection{K-means-ARIMA}
In \cite{kmeansANN} authors combined k-means clustering with autoregressive neural networks for hourly solar forecasts. Similarly, authors in \cite{ARIMAann} used wavelet packet transform along with ANN for the application of wind power forecasting.

However, aforementioned hybrid methods posed a common issue of escalating the computational complexity. 
In addition, if one method in the hybrid combination has the poor performance, it is going to affect the overall hybrid method. However, by including an appropriate optimization algorithm to optimize model parameters, the efficacy and performance can be improved. Thus, maintaining a trade-off between the computational complexity and performance is of utmost importance for the hybrid approach.
\section{Data pre-processing techniques for energy forecasting}\label{sec:preprocessing}
Data representation and pre-processing is one of the early key stages in data analytics. Processing raw data with irrelevant and redundant information can produce misleading and incorrect forecasting results. Data pre-processing involves various methods such as normalization, cleaning, transformation, feature engineering, dimensionality reduction, etc. It transforms raw data to a final training dataset which can be further fed to the data processing techniques \cite{zhu2016short}.
For energy systems, various authors have emphasized the use of different pre-processing techniques to improve the forecasting accuracy, as discussed in the following subsections.
\subsection{Singular value decomposition (SVD)}
Dimensionality reduction and feature engineering are two highly regarded pre-processing techniques. Kaur \textit{et. al.} in \cite{kaur2019smart} and \cite{kaur2018tensor} used SVD as the base technique to decompose high dimensional data into lower dimensions prior to energy forecasting. The authors employed tensors and SVD matrix decomposition to achieve dimensionality reduction. 
Similarly, authors in \cite{wang2016sparse} presented a data decomposition and compression method for user load profiles based on k-means clustering and SVD (k-SVD) method. Firstly, a sparse coding technique is used over load profiles to compress the data by exploiting the sparsity present in load profiles. Then, k-SVD is proposed to decompose and extract the partial usage patterns (PUPs) in it. They claimed that their technique outperforms other existing data compression methods namely discrete wavelet transform (DWT) and standalone k-means clustering.
\subsection{Principal component analysis (PCA)} 
PCA is a classical dimensionality reduction technique used to transform the higher data dimensions into lower representations in the form of orthogonal matrix known as principal components. It is proposed in \cite{huang2019optimization} to eliminate the features from a PV generation dataset that has little or no significance in the forecasting output. It allows the proposed method to make  more precise predictions. Authors compared their results with differential evolution (DE) and particle swarm optimization (PSO) and observed least RMSE values using their method. 
\subsection{Auto-encoders (AE)}
AEs are type of neural networks used to encode high dimensional data into lower representations using the encoding layer. The performance of AEs is checked with the help of decoding layer by monitoring the reconstruction ratio. Furthermore, there exist different variants of AEs such as sparse auto-encoders (SAE). A special class of SAE is used in \cite{chen2016framework} as a framework to identify the over-voltage to classify the faults and determine the power quality disturbances.
Furthermore, AEs are used in \cite{zhang2020data} with LSTM to improve PV  estimation accuracy in variable weather conditions. The authors claimed to reduce the impact of weather uncertainty using encode-decoder framework for day-ahead PV forecasting for multiple locations.
\subsection{Convolutional auto-encoders (CAE)}
AEs incorporated with the convolutional layers of CNN are termed as CAEs.
In \cite{ryu2019convolutional}, authors emphasized the role of data compression and dimensionality reduction using CAEs from the perspective of data storage and transmission. They proposed a feature extraction technique based on CAEs to capture daily and seasonal variations by representing $8640$ dimensional space into $100$-dimensional vector. The authors reported $19$-$40$\% decline in the reconstruction error using their technique and a $130$\% increase in compression ratio as compared to standard methods.
Furthermore, Shao \textit{et. al.} used CAE integrated with LSTM to perform multi-step STLF involving energy, time, and user behavior components. Firstly, they combined CAE and k-means to identify user behavior and then employed LSTM to train time and energy components for STLF with $10$\% improvement in overall prediction performance \cite{shao2020multi}.
\subsection{Variational auto-encoders (VAE)}
VAEs are relatively a new class of AEs in energy systems which employ the concept of variational inference and Bayesian optimization to carry out the encoding and decoding operations \cite{Zhang2018,ibrahim2021variational}. VAE integrated with recurrent neural layers and its variants have been recently explored for the application of anomaly detection and to make the forecasting process more efficient \cite{pereira2018unsupervised,Dairi:2020,bachhav2019latent,wang2020variational}. To reduce the dimensionality arising from time lags considered in renewable generation forecasting, VAEs have been implemented in \cite{kaur2021ICC} before performing forecasts with BLSTM.  
VAE combined with Bayesian BLSTM has been employed in \cite{kaur2021vae} to carry out dimensionality reduction of model parameters and forecasting respectively for solar generation dataset. The authors reported lower reconstruction and RMSE errors using their technique along with uncertainty quantification in model parameters of BLSTM. 
\section{Case Study}\label{sec:casestudy}
This section presents a case study for comparative analysis of various forecasting methods discussed in above sections using two different datasets. To be specific, RNN and LSTM are chosen as benchmark DL models due to their superior performance reported in existing papers. For similar reasons, ARIMA, SVR, Bayesian ANN, and BLSTM are chosen as representatives of commonly used statistical, ML, and PDL methods, respectively.
\subsection{Description of data}
In order to evaluate the performance of forecasting methods, two energy datasets from the AEP \cite{aep} and Victorian energy consumption benchmark \cite{vic} are taken. AEP dataset consists of hourly energy consumption values from October $2004$ to August $2018$. Victorian dataset includes half hourly energy consumption values from April $2012$ to March $2014$ for $25$ Victorian houses. To implement point forecasting methods, $5$ time lags (equivalent to $5$ hours for AEP dataset and $2.5$ hours for Victorian dataset) are considered. For PDL methods, $24$ time lags (equivalent to $24$ hours for AEP dataset and $12$ hours for Victorian dataset) are considered. 
Results are implemented in python environment using Keras and Tensorflow libraries.
\subsection{Evaluation metrics}
RMSE and MAE are the most commonly used evaluation metrics to evaluate the point forecasting methods \cite{toubeau2018deep},\cite{sun2019using}.
The RMSE represented by {$\epsilon$} and MAE represented by $\rho$ are defined using following equations:
\begin{equation}
\epsilon=\sqrt{\frac{1}{n} \sum\limits_{t=1}^{n}(Y_{pred}(t)-Y_{act}(t))^2}
\label{rmse}
\end{equation}
\begin{equation}
	\rho=\frac{1}{n}\sum\limits_{t=1}^{n}\mid{{Y_{pre(t)}-Y_{act(t)}}}\mid
	\label{mape}
\end{equation}
where $Y_{pred}(t)$ and $Y_{act}(t)$ are the predicted and actual values at time stamp $t$ and $n$ is the total number of samples, respectively. Thus, the main objective for forecasting model is to minimize the error in predicted values as:
\begin{equation}
\min_{\mathcal{\theta}} \epsilon,\rho
\end{equation}
where $\theta$ represents the model parameters for each forecasting method such as weights and biases or lag coefficients.
Furthermore, average Pinball loss is an important evaluation metric for probabilistic methods \cite{yang2019bayesian,sun2019using}. 
So, to evaluate Bayesian ANN and Bayesian BLSTM, average Pinball loss is computed focusing on the sharpness and consistency of the approximated distribution.
In this regard, least value of Pinball loss is more desirable. 
For actual data values ($Y_{act}$) and predictions at $t^{th}$ time-stamp (${\hat{Y}_{t,q}}$), Pinball loss
over percentile $q$ $\in[0,1]$ is formulated as:
\begin{equation}\label{eq:Pinball}
Pinball(Y_{act},\hat{Y}_{t,q},q)=\begin{cases}
(Y_{act}-{\hat{Y}_{t,q}}) q & Y_{t} \ge {\hat{Y}_{t,q}}\\
({\hat{Y}_{t,q}}-Y_{act}) (1-q) & Y_{act} \le {\hat{Y}_{t,q}}
\end{cases}  
\end{equation}
Furthermore, for confidence $(1-\gamma)$ $\times$ $100$, Winkler score is computed using:
\begin{equation}
Winkler=\begin{cases}
\delta & lb_t \le y_t \ge ub_t\\
\delta + 2(lb_t-y_t) / \gamma & lb_t \ge y_t\\
\delta + 2 (y_t-ub_t) / \gamma & ub_t \le y_t\\
\end{cases}
\end{equation}
where $lb_t$ and $ub_t$ represent the lower and upper bounds of probabilistic forecasts at interval $t$, respectively. And, $\delta$ = $ub_t -lb_t$ is the PI width at $t$.
\subsection{Comparative analysis}
\begin{table*}
	\renewcommand{\arraystretch}{1.5}
	\caption{Comparative analysis on evaluation metrics for Victorian and AEP case study}
	\begin{center}
		Bold values represent the best performing methods in the form of least errors for each case of dataset.
	\end{center}
	\begin{center}
		\begin{tabular}{|p{8mm}|p{22mm}|p{12mm}|p{12mm}|p{12mm}|p{12mm}|p{12mm}|p{12mm}|p{12mm}|p{12mm}|}
			\hline
			\multicolumn{2}{|c} {} & \multicolumn{4}{c|} {Victorian energy consumption dataset} &\multicolumn{4}{c|}{AEP dataset}\\	
			\hline
			
			\multicolumn{2}{|c|} {} & \multicolumn{2}{c|} { Two years}	
			& \multicolumn{2}{c|}{One year}
			& \multicolumn{2}{c|} {Ten years} & \multicolumn{2}{c|}{One year}\\
			\hline
			
			Sr. no. & Method & RMSE & MAE & RMSE & MAE  & RMSE& MAE&RMSE&MAE\\
			\hline
			\hline
			1&	ARIMA & - & - & $0.0753$ & $0.0566$  & -  &-& $0.0429$ & $0.0370$\\	
			
			2&SVR-rbf& $0.0813$ & $0.0795$  &$0.0792$ &$0.0605$  &$0.0363$& $0.0298$&$0.0415$ &$0.0331$\\
			
			3&SVR-linear& $0.0676$  & $0.0651$  & $0.0749$   & $0.0562$ &$0.0310$ &$0.0242$ &$0.0362$ &$0.0285$ \\
			
			4&RNN-tanh  & $0.0736$ &  $ 0.0559$ &   $0.0774$  &  $0.0579$ &
			$0.0193$ & $0.0141$ &$\mathbf{0.0371}$&$\mathbf{0.0285}$\\
			
			5&RNN-relu & $\mathbf{0.0631}$ &  $\mathbf{ 0.0463}$&$\mathbf{0.0701}$ & $\mathbf{0.0519}$ &
			$0.0290$ & $0.0228$ &
			$0.0527$ & $ 0.0424$\\
			
			6&LSTM-tanh &$0.0805$ & $0.0608$ &$0.0851$ &  $0.0637$ 
			&$\mathbf{0.0189}$&$\mathbf{0.0136}$& $0.0472$ & $0.0368$ \\
			
			7&LSTM-relu &$0.0689$  &  $0.0516$ &$0.0804$ & $0.0599$& $0.0239$&$0.0183$ & $0.0551$&$0.0430$\\
			
			8&CNN  & $0.0665$&   $0.0495$   &$0.0796$ &$0.0597$ &$0.0197$ &$0.0142$ &$0.0445$ &$0.0342$\\ 
			9&Bayesian ANN &$0.0481$&$0.0417$&$0.0438$&$0.0405$&$0.0198$&$0.0154$&$0.0564$&$0.0445$\\
			
			10&Bayesian BLSTM &$\mathbf{0.0251}$ &$\mathbf{0.0281}$&$\mathbf{0.0305}$&$\mathbf{0.0260}$&$\mathbf{0.0168}$&$\mathbf{0.0126}$&$\mathbf{0.0167}$&$ \mathbf{0.0121}$\\
			\hline 
		\end{tabular}
	\end{center}
	\label{main}
\end{table*}
\begin{table*}[!htb]
	\centering
	\renewcommand{\arraystretch}{1.5}
	\caption{Pinball score by PDL forecasting methods }
	\begin{tabular}{p{10mm}p{30mm}p{25mm}p{25mm}p{25mm}p{25mm}}
		\hline
		Sr. no.&	Method &Victorian (Two years)&Victorian (One year)&AEP (Ten years) &	AEP (One year)\\
		\hline	
		\hline
		1&Bayesian ANN&$0.0019$&$0.0210$&$0.0050$ &$0.0150$\\
		2&Bayesian BLSTM&$0.0016$&$0.0180$&$0.0030$&$0.0070$\\
		\hline
	\end{tabular}
	\label{Pinball}
\end{table*}
\begin{table*}[!htb]
	\centering
	\renewcommand{\arraystretch}{1.5}
	\caption{Winkler score by PDL forecasting methods}
	\begin{tabular}{p{10mm}p{30mm}p{25mm}p{25mm}p{25mm}p{25mm}}
		\hline
		Sr. no. &	Method &Victorian (Two years)&	Victorian (One year) &AEP (Ten years) &	AEP (One year)\\
		\hline	
		\hline
		1&	Bayesian ANN&$0.7264$&$0.7834$&$0.1273$&$0.2341$\\
		2&	Bayesian BLSTM&$0.5432$&$0.6865$&$0.0703$&$0.0902$\\
		\hline
	\end{tabular}
	\label{winkler}
\end{table*}
\begin{figure*}[ht!]
	\centering
	\subfigure[Victorian dataset for different forecasting methods] 
	{\includegraphics[height=0.2\textheight,width=0.45\textwidth]{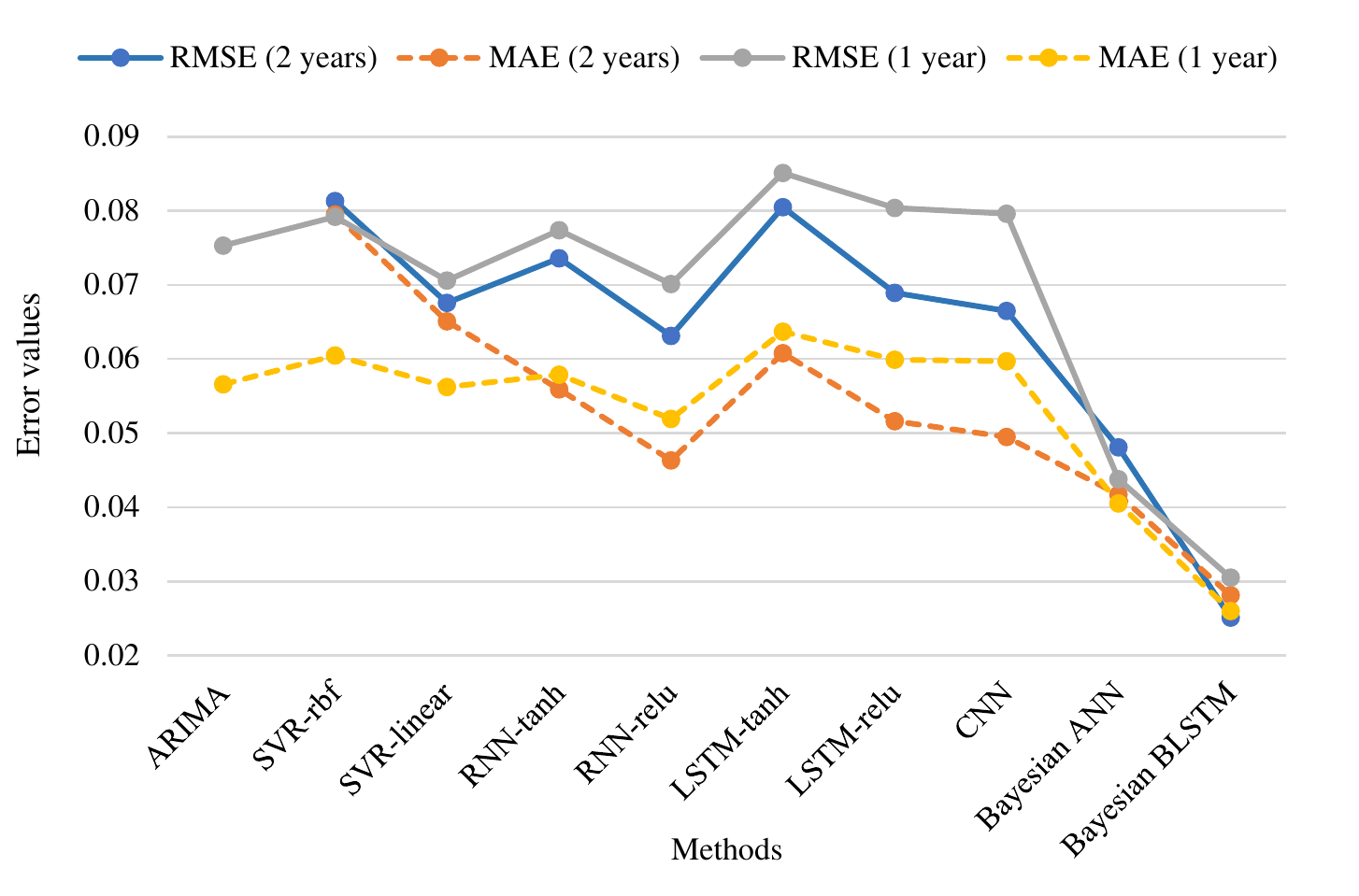}}
	\subfigure[AEP dataset for different forecasting methods]
	{\includegraphics[height=0.2\textheight,width=0.45\textwidth]{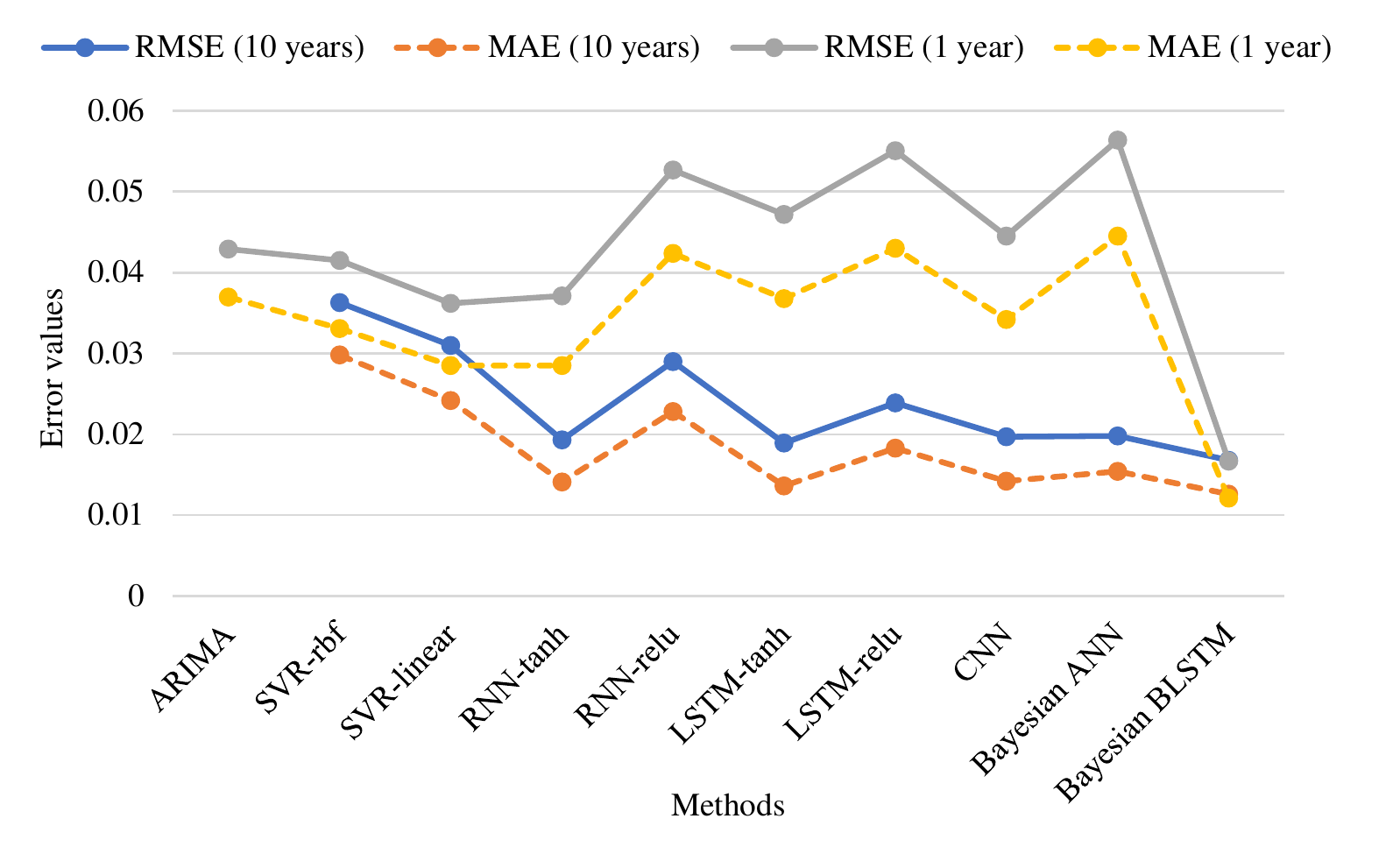}}
	\caption{Comparative analysis of forecasting methods based on evaluation metrics for different datasets with different sample sizes.}
	\label{error}
\end{figure*}
As shown in Table \ref{main}, both datasets are further divided into two cases each to evaluate the impact of different sample sizes. 
For the Victorian energy, two and one year of consumption sample size are considered. While for AEP, ten and one year of consumption data are considered. 
Then, ARIMA, SVR, RNN, LSTM, CNN, Bayesian ANN, and Bayesian BLSTM are trained with $70\%$ of the total samples for each case and tested for remaining 30\%.
For DL methods, $40$ neural units per layer, $100$ epochs, and batch size of $1000$ are chosen. $Adam$ is used as an optimizer to minimize the loss function of mean squared error (MSE). Additionally, RNN and LSTM methods are considered with two different variations in the form of activation functions namely, hyperbolic tangent function (\textit{tanh}), and  rectified linear unit (\textit{Relu}). Activation functions play an important role in activating the hidden layer in neural networks. Also, SVR is modeled with radial basis function (\textit{rbf}) and \textit{linear} kernel. Then, RMSE and MAE values are computed on the testing dataset using (1) and (2), respectively. 

\begin{figure*}[!h]
	\centering
	\subfigure[Victorian energy consumption dataset]
	{\includegraphics[height=0.22\textheight,width=0.45\textwidth]{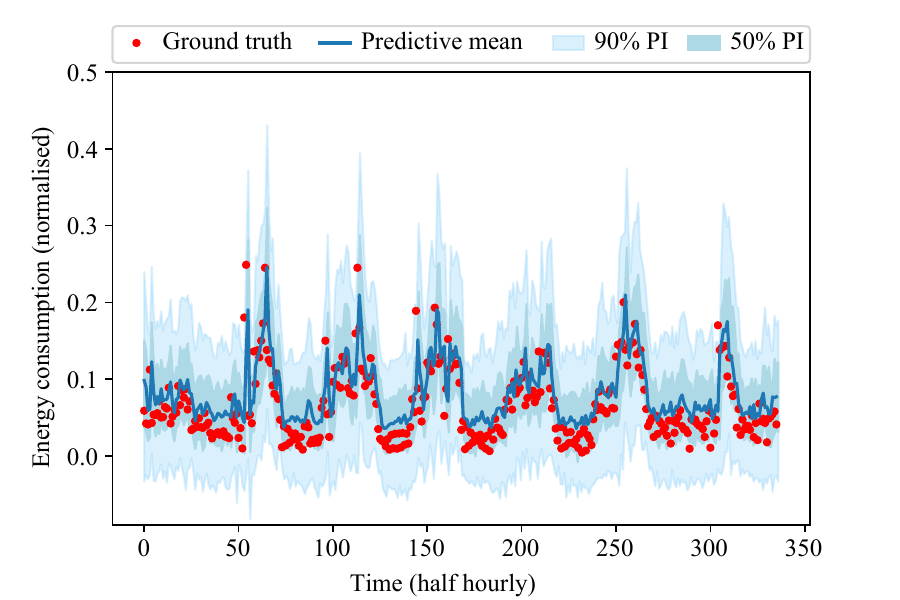}}
	\subfigure[AEP dataset]
	{\includegraphics[height=0.22\textheight,width=0.45\textwidth]{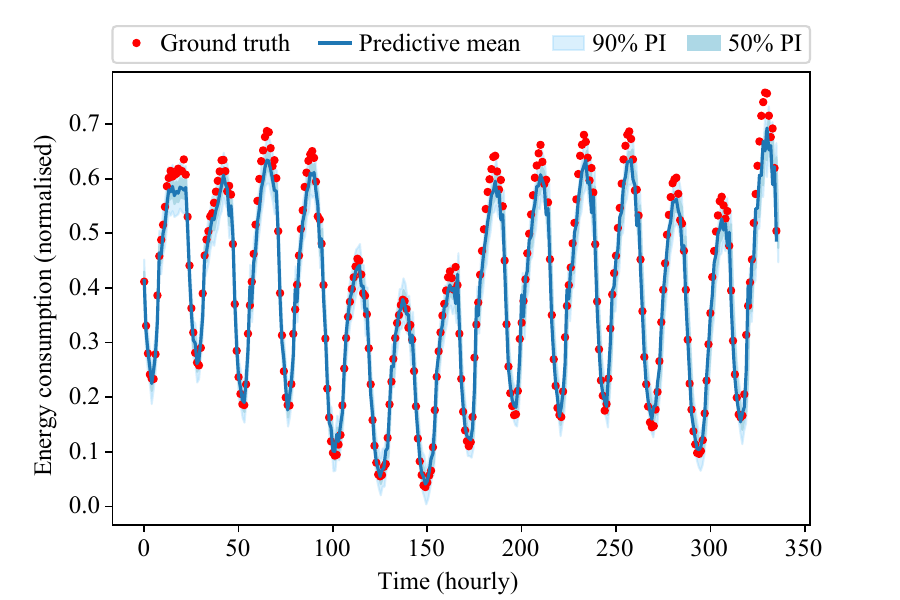}}
	\caption{Prediction intervals (PIs) with ground truth and predictive mean using Bayesian BLSTM}
	\label{BLSTM}
\end{figure*}

As demonstrated in the table, Bayesian BLSTM performs similar to LSTM (with \textit{tanh}) for larger training datasets and stationary time-series, i.e., with 10 years of data for AEP. 
However, for non-stationary and smaller training samples, Bayesian BLSTM provides the best results by achieving the least error value, as reflected in Table \ref{main}. It should be noted that 
from point forecasting methods, RNN with \textit{Relu} gives second best results in terms of lower error values.
In addition, Bayesian BLSTM provides PIs for future predictions and forecasts with more accuracy in relation to the ground truth, as reflected in Fig. \ref{BLSTM}.
RNN and LSTM have less error as compared to the statistical ARIMA and traditional ML methods such as SVR. For AEP dataset, LSTM is performing better with larger samples. For smaller sample size, RNN demonstrates less error and computational cost. Models with \textit{tanh} function perform better with uniform data, such as the AEP dataset. However, with higher variability and seasonal trends, i.e, for the Victorian dataset, \textit{Relu} performs better as it sparsely activates neurons rather than activating all neurons at the same time. 
Furthermore, datasets with more samples tend to train better and yield greater accuracy. In addition, ARIMA takes a longer time to train for larger datasets and thus, fails to generate the output. SVR with \textit{linear} kernel has less error compared to SVR with \textit{rbf} kernel. When dataset has linearly separable features, \textit{linear} kernel will be sufficient to achieve superior performance.

Tables \ref{Pinball} and \ref{winkler} display the Pinball and Winkler scores, respectively given by Bayesian ANN and Bayesian BLSTM for different data sizes. It can be inferred from the values that in case of more data and hence larger training set, scores are lower indicating the improvement in accuracy. Also, Bayesian BLSTM outperforms Bayesian ANN with the help of extensive training capability of bidirectional layer.

Fig. \ref{error} reflects the comparison graphs of errors observed by forecasting methods considered in Table \ref{main} with respect to each sample size. It can be inferred from the figures that for point forecasting, standard DL methods provide least error irrespective of the data size and variability. 
However for PDL forecasting, Bayesian BLSTM outperforms all the other comparative methods and generate PIs.

\begin{figure*}[!ht]
	\centering
	\subfigure[Victorian energy consumption dataset]
	{\includegraphics[height=0.22\textheight,width=0.45\textwidth]{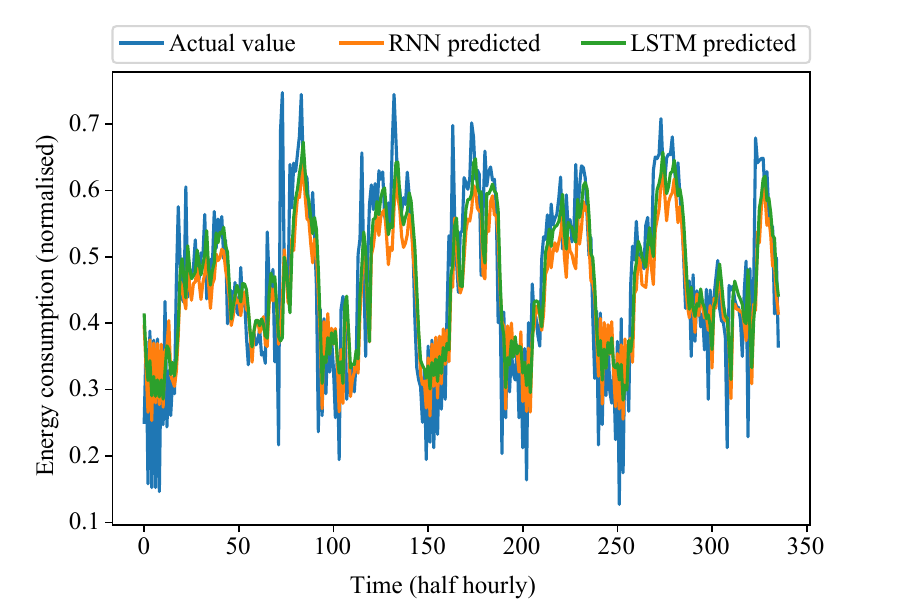}}
	\subfigure[AEP dataset]
	{\includegraphics[height=0.22\textheight,width=0.45\textwidth]{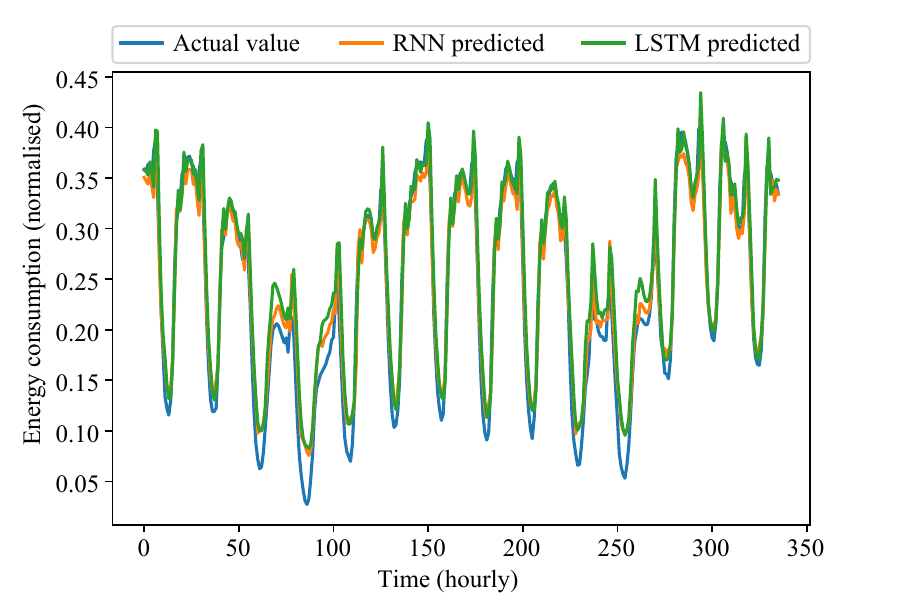}}
	\caption{Predicted vs actual values for RNN and LSTM with one week and two weeks of test data for the Victorian and AEP dataset, respectively.}
	\label{rnnlstm}
\end{figure*}
Furthermore, Fig. \ref{rnnlstm} shows the actual and predicted energy consumption values using RNN and LSTM on each dataset. The results are plotted for activation functions which produce superior performance for a given algorithm. Normalized energy consumption values are plotted over half hourly and per hour intervals for Victorian and AEP dataset, respectively. From Fig. \ref{rnnlstm}(a), it is inferred that the Victorian dataset has higher variability leading to more uncertainty in the energy consumption and hence, the performance of the models is slightly poorer. To deal with its non-stationary nature, logarithmic transformation is adapted. However, in Fig. \ref{rnnlstm}(b), AEP dataset is uniformly distributed for which predicted values show greater similarity with actual values and thus, less errors.
Furthermore, Fig. \ref{BLSTM} represents a comparison graph of actual energy consumption values (ground truth) with predictive mean over $90\%$ and $50\%$ prediction PIs generated by Bayesian BLSTM on both the datasets. 
PIs reflect the intervals for future probabilities on different percentiles which enable the PDL methods to quantify for uncertainties.
Fig \ref{BLSTM}(a) clearly shows wider PIs as compared to \ref{BLSTM}(b) as there is more variability in Victorian consumption values than AEP.
\begin{figure*}[!ht]
	\centering
	\subfigure[Victorian energy consumption dataset] {\includegraphics[height=0.28\textheight,width=0.8\textwidth]{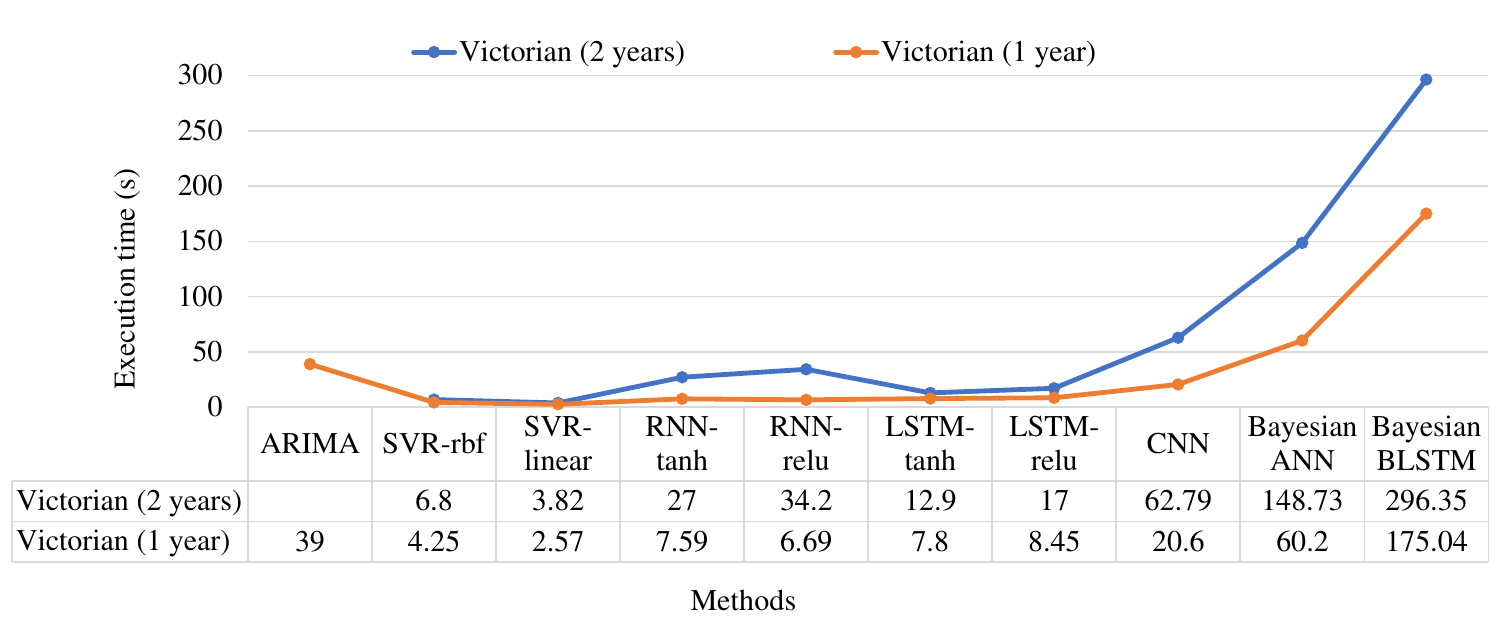}}
	\subfigure[AEP dataset]
	{\includegraphics[height=0.28\textheight,width=0.8\textwidth]{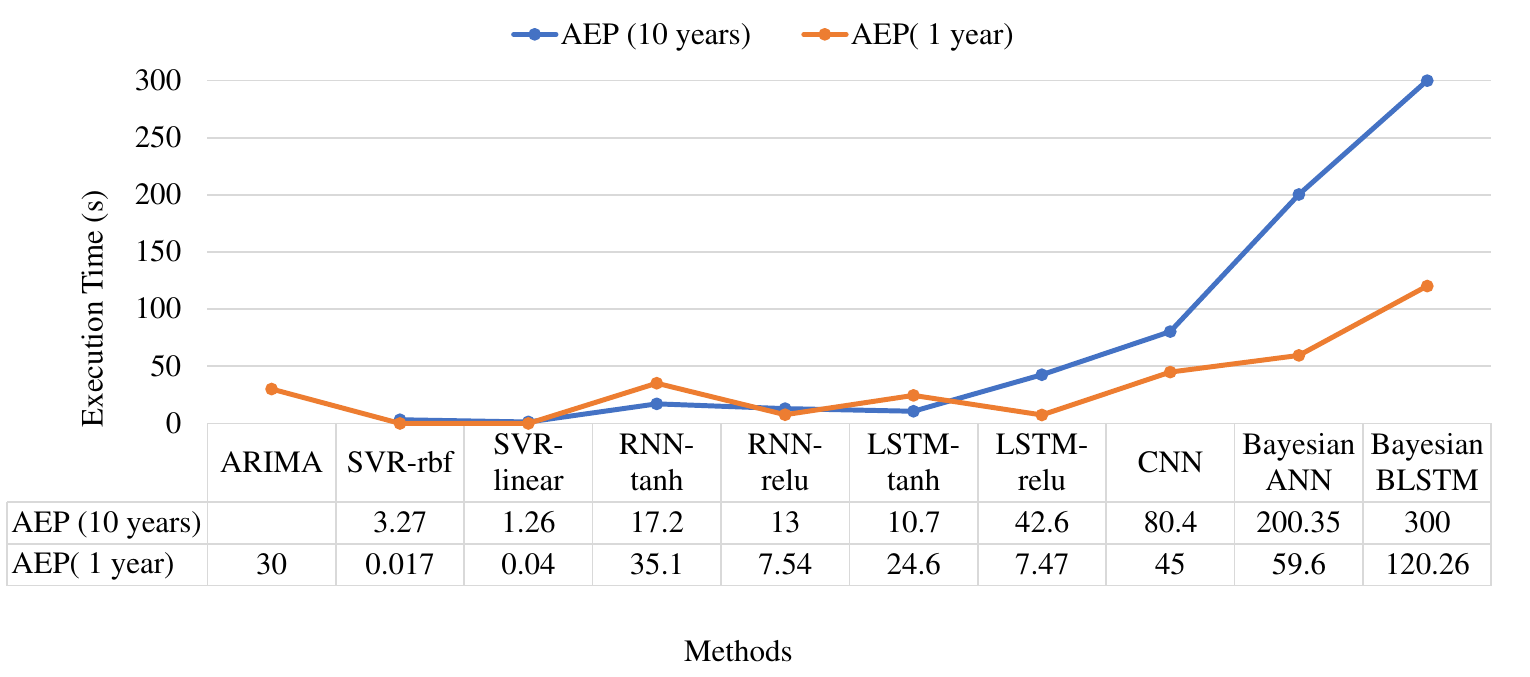}}
	\caption{Comparative analysis of Execution time for respective forecasting method}
	\label{time}
\end{figure*}

In addition, a comparative plot for the execution time taken by each method is presented in Fig. \ref{time}. Note that, the reported execution time precisely belongs to the training process of each forecasting method. Time taken for error computation and parameter tuning is not included. As shown in the figure, PDL methods take more time compared to point forecasting methods. There is trade-off involved between the time complexity and forecasting accuracy for the PDL methods, which needs to be maintained.
\section{Conclusion and future directions}\label{sec:conclusion}
In this paper, a comprehensive review of classical and advanced forecasting methods is presented for modern energy systems. A number of statistical, AI-based, probabilistic, and hybrid methods are discussed in detail with respect to their applications in energy systems. In addition, impact of data pre-preprocessing techniques on forecasting performance is also highlighted.
After conducting a comparative case study on two different datasets, it is inferred that DL methods with appropriate activation function and hyper-parameter tuning yield higher forecasting accuracy than the traditional statistical and ML methods. However, uncertainty in the energy data from exogenous factors is a major challenge that can be tackled more efficiently with probabilistic methods. To support this,
we implemented Bayesian ANN and Bayesian BLSTM as PDL forecasting techniques and the latter outperformed all the other comparative methods by demonstrating least error values. However, it exhibit high computational cost in the terms of time complexity and processing units. So, probabilistic forecasting techniques need to be explored more in the future work and potential solutions to reduce computational cost need to be proposed in the energy domain.
Moreover, forecasting multi-dimensional demand and generation while ensuring high accuracy is another issue that requires substantial attention.
\bibliographystyle{iet}

\end{document}